\pdfoutput=1

\documentclass[11pt]{article}

\usepackage[preprint]{acl}

\usepackage{times}
\usepackage{latexsym}

\usepackage[T1]{fontenc}

\usepackage[utf8]{inputenc}

\usepackage{microtype}

\usepackage{inconsolata}

\usepackage{graphicx}
\usepackage{enumitem}
\usepackage{soul}

\usepackage{listings}
\newcommand{\lp}[1]{\textcolor{black}{#1}}
\newcommand{\fm}[1]{\textcolor{black}{#1}}
\newcommand{\bb}[1]{\textcolor{black}{#1}}
\newcommand{\LB}[1]{\textcolor{black}{#1}}
\newcommand{\fp}[1]{\textcolor{black}{#1}}

\usepackage{adjustbox}
\usepackage{rotating}
\usepackage{langsci-gb4e}
\title{\fm{Constraining constructions with WordNet: pros and cons for the semantic annotation of fillers in the Italian Constructicon}}

\author{Flavio Pisciotta \\
  University of Salerno \\
  \texttt{fpisciotta@unisa.it} \\\And
  Ludovica Pannitto \\
  University of Bologna \\
  \texttt{ludovica.pannitto@unibo.it} \\\And
  Lucia Busso \\
  Aston University \\
  \texttt{l.busso@aston.ac.uk} \\\AND
  Beatrice Bernasconi \\
  University of Turin \\ 
  \texttt{beatrice.bernasconi@unito.it} \\\And
  Francesca Masini \\
University of Bologna \\
\texttt{francesca.masini@unibo.it}
 }
  
\begin{document}

\maketitle
\begin{abstract}
\fm{The paper discusses the role of WordNet\lp{-based semantic classification} in the formalization of constructions, and more specifically in the semantic annotation of \lp{schematic} fillers, in the Italian Constructicon.} \LB{We outline how the Italian Constructicon project uses Open Multilingual WordNet topics to represent semantic features and constraints of constructions.}
\end{abstract}

\section{Introduction}

\begin{figure*} 
\small{
\begin{lstlisting}
#cxn-id = 171
#cxn = fare Npsych
#function = cause to feel ref:B 

ID  UD.FORM     LEMMA   UPOS    FEATS       HEAD    DEPREL
A   _           fare    VERB    _           0       root
B   _           _       NOUN    Number=Sing A       obj

REQUIRED    WITHOUT             SEM.FEATS           ADJACENCY   IDENTITY
1           _                   _                   _           _
1           CHILDREN:DEPREL=det OntoClass=feeling   _           _
\end{lstlisting}
}

\captionof{lstlisting}{Example of CoNLL-C annotation for the light verb cxn \textit{fare} N\textsubscript{feeling} `make feel N\textsubscript{feeling}' (lit. do N\textsubscript{feeling}) \cite{pisciotta-masini}. \LB{Since this construction only occurs with a psychological noun in the singular form, the features of the noun are specified with "number=sing", and the semantic layer uses the topic of "feeling" to constraint the nouns that can occurr in the second slot of the construction. }}
\label{fig:conll-c}
\end{figure*}

In Construction Grammar (CxG, \citeauthor{hoffman&trousdale2013}, \citeyear{hoffman&trousdale2013}), the basic units of linguistic description are constructions (cxns), which are conventionalized pairings of form and function~\cite{goldberg1995constructions, goldberg2006constructions}. Crucially, cxns can vary in complexity and schematicity, including not only words, but also more complex and/or abstract units such as predicative structures, idioms, and word formation processes. CxG, as other usage-based approaches, assume that cxns are not stored as a mere list, but as a structured network (the \textit{Construct-i-con}) in which cxns are linked by different kinds of relationships~\citep{diessel2019, diessel2023}. 

\LB{Despite traditional research in CxG not focusing much on language as a system, recent years have seen a growing interest in \textit{Constructicography}, a blend of ``Practical Lexicography'' and CxG~\citep{boasetal2019}. That is, the notion of ``Constructicon'' has acquired an additional meaning. Beside relating to the structured inventory of all constructions in a language, it has come to indicate a linguistic resource that aims at representing and formalising the network of constructions in a given language~\cite{lyngfelt2018constructicography}.  }

Constructicography, therefore, is the research field that aims to build Constructicons, that is, to develop repositories of cxns that consistently and coherently describe the grammar (and thus the constructional network) of a specific language. Constructicons already exist for a number of languages (e.g., \citealt{jandaetal2018,lyngfeltetal2018,torrentetal2018}), and are often linked to the FrameNet enterprise~\citep{bakeretal1998}. In fact, Frame Semantics is considered a ``sister'' framework to CxG - as both theories stem from Fillmore's work on semantic roles~\cite{fillmore1968case} - and is typically used to represent semantic aspects of constructions~\cite{borinframenets}.

Despite Constructicography being a fast-growing research area in many languages, Italian has so far been at the periphery of it. Not only there is no Constructicon, but there is also no published Italian FrameNet, although there have been several attempts at developing such a resource~\cite{tonellietal2009,lencietal2010,basili2017}.
The present contribution introduces the Italian Constructicon (ItCon) project \citep{masini-etal-2024-adoc}. This project aims to bridge this gap by building an open and collaborative resource \LB{that is designed to be interoperable with existing resources for Italian (treebanks, lexical databases, corpora).}
\LB{Crucially, we outline how we use WordNet-based semantic classification to represent the semantic layer of Italian constructions.}

As it stands, the resource is still in its infancy.
Therefore, the primary goal so far is to develop a solid theoretical and operational background for the project. In this contribution, we focus specifically on how to constrain the generative power of cxns with respect to the semantic productivity of the open slots of semi-specified cxns~\citep{suttle&goldberg2011, perek2016}, and how this problem can be addressed operationally through the integration of data from WordNet(s) available for Italian~\citep{italwordnet2000, multiwordnet2002} in our annotation format. We will proceed by briefly describing the architecture of ItCon and the annotation format of constructional entries (Section~\ref{sec:architecture}), and then we discuss how our annotation scheme can benefit from the connection with WordNet, as well as the possible limitations of such proposals (Sections~\ref{sec:omw} and~\ref{sec:future}).

\section{Architecture of the Italian Constructicon}
\label{sec:architecture}
\vspace{-1mm}
ItCon consists of three linked structures:
\begin{itemize}[nosep]
    \item a \textbf{database of cxns};
    \item the \textbf{graph of cxns}, \lp{where each node represents a cxn in terms of the set of constraints that it expresses and edges represent horizontal and vertical links holding between cxns;}
    \item a body of \textbf{annotated examples} \lp{in CoNLL-U format \cite{nivre2016universal}}, incrementally built by annotating instances of a specific cxn (i.e., constructs) in texts by means of a specific feature in the \texttt{MISC} field.
\end{itemize}

In the \textbf{database of cxns}, each entry describes a cxn through a number of text fields and tags. They serve the purpose of specifying information about the properties and behavior of the constructs, as well as linking the database entry to a node in the \textbf{graph of cxns} and to a subset of the \textbf{annotated examples}. 

Each node in the graph of cxns \bb{consists of a \lp{columnar}  
formalization customized for cxns representation\lp{, based on CoNLL-X format~\cite{buchholz2006conll}} and therefore named CoNLL-C \citep{masini-etal-2024-adoc},} 
that can be converted into a Grew query~\citep{guillaume2021} in order to match occurrences of the cxn in CoNLL-U annotated corpora, \fp{ i.e., Universal Dependencies (UD,~\citealt{nivre2016universal}) treebanks}. The generative power of the cxn gets constrained at this level, as it is necessary to narrow down the possible set of matched occurrences. This is done through a set of fields specifying formal and functional constraints, which we briefly describe.

\subsection{The CoNLL-C format}

The CoNLL-C format is a UD compatible format\footnote{For a comprehensive description of the format and the relevant fields, see \cite{annotating-cxns}.}. As shown in \lp{Listing} \ref{fig:conll-c}
\footnote{The columnar format 
was split in two lines for space reasons.}, each formalized cxn is described by a set of metadata (i.e., the 
lines \lp{prefixed by $\#$}
) that specify holistic properties of the cxn (such as its \textit{semantic function}), and by a number of fields, containing a token-by-token description \lp{of the cxn components}. The first 7 fields (\texttt{ID}, \texttt{UD.FORM}, \texttt{LEMMA}, \texttt{UPOS}, \texttt{FEATS}, \texttt{HEAD}, \texttt{DEPREL}) can be mapped on 
the \lp{matching} fields
in CoNLL-U format. Since one of the aims of such formalization is to match the relevant constructs in UD-annotated corpora, some other fields were added to \lp{formally} constrain the queried pattern. 
\lp{They include information such as whether a token is necessarily expressed (\texttt{REQUIRED}), the possibility of intervening material within the cxn (\texttt{ADJACENCY}), any excluded values (\texttt{WITHOUT}), as well as the need for sharing of some features between two tokens (\texttt{IDENTITY}).}


Taking into account the aforementioned fields, the formalization in \lp{Listing \ref{fig:conll-c}} can be rewritten in Grew query language~\citep{guillaume2021} as follows:

\lstset{basicstyle = \fontsize{10}{12}\selectfont}
\begin{lstlisting}
pattern {X1 [lemma='fare']; 
        X2 [upos=NOUN, Number=Sing];
        X1 < X2; 
        X1 -[obj]-> X2}
without {X2 -[det]-> X3}
\end{lstlisting}


However, such formalization can only partially constrain the set of matching patterns. For instance, by searching the PoSTWITA-UD treebank~\citep{sanguinetti-etal-2018-postwita} applying such a query, we obtain both patterns corresponding to \textit{fare} N\textsubscript{feeling} `make feel N\textsubscript{feeling}' cxn (\ref{ex:true}), as well as false positives (\ref{ex:false}):

\begin{exe}
\ex \label{ex:true} \textit{fare schifo} `to disgust', \textit{fare paura} `to frighten', \textit{fare piacere} `to please'
\ex \label{ex:false} \textit{fare demagogia} `to be demagogic', \textit{fare parte} `to be part', \textit{fare cassa} `to make profit'
\end{exe}


Patterns in (\ref{ex:false}) are not instances of the cxn we want to match: they do not express a causative nor a psychological semantics (since they do not involve nouns expressing psychological states). For such reasons, we added the \texttt{SEM.FEATS} field, where semantic features of the tokens filling the empty slots can be specified. 

As for now, the semantic features include the semantic class (\texttt{OntoClass}) for nouns and verbs, and \bb{A}ktionsart (\texttt{Aktionsart}) for verbs only. Given the need for interoperability with other resources, however, cross-linguistically and cross-resource shared annotation schemes are needed for such features. In the following section, we show how we intend to employ WordNet data to annotate the \lp{\texttt{OntoClass} semantic feature }
in our cxns, discussing the advantages and limitations of such an approach.

\section{WordNet for semantic classification}
\label{sec:omw}

As mentioned, one of the semantic features we included in the formalization is \texttt{OntoClass}. In this category, we annotate the semantic classes of \fm{slots in our }cxns using Open Multilingual WordNet (OMW) topics~\citep{bond-foster-2013-linking}, as currently mapped onto Italian MultiWordNet~\citep{multiwordnet2002}. These topics are the Lexicographer files used by Princeton WordNet~\citep{wordnet}, and correspond to the top nodes used to build the hierarchy of the four WordNet categories: noun, verb, adjective, and adverb~\cite{miller-et-al-1990}. Currently, we decided to employ the tagset only for nouns (26 classes) and verbs (15 classes).

We chose to employ OMW topics over developing an original classification for several reasons. Firstly, using OMW topics provides ItCon with an annotation scheme that is cross-linguistically  interoperable, and a shared standard. Even though at the moment of writing (January 2025) no other Constructicon annotates semantic constraints on fillers of cxns, we hope that in the future using OMW will provide an easy and theoretically-grounded way to link constructicons.  

Secondly, OMW topics have been already used as a semantic classification in sense-tagged corpora\footnote{See, for instance, SemCor~\citep{miller-etal-1994} and the subsequent work on multilingually aligning sense-tagged corpora~\citep{bentivogli-pianta, attardi-etal-2010}.}, which potentially makes ItCon interoperable with other, not CxG-related sense-tagged or WordNet-related resources.

Another advantage of using OMW's ontology is that it includes the hierarchy of synsets, which allows \bb{for} flexibility in determining the level of granularity needed in tagging semantic constraints case by case, while still relying on a relatively small number of tags\footnote{The lower number of tags is the reason why we chose OMW topics over EuroWordNet~\citep{Rodríguez1998} top nodes (n = 63).}.

As for now, \fm{we found ourselves resorting to such \bb{a} semantic classification in constraining the matching process of our cxns, although a more systematic testing is necessary to prove its usefulness.} 
For instance, by tagging the noun slot in the \textit{fare} N\textsubscript{feeling} cxn with the class \texttt{noun.feeling} (Listing \ref{fig:conll-c}), we are able to exclude most of the false positives in the matching process (cf. \ref{ex:true}-\ref{ex:false}): 

\begin{exe}
\ex \label{ex:falsetag}Instances of \textit{fare} N\textsubscript{feeling}:
    \begin{xlist}
    \ex \label{ex:truetag1}
    \glll\textit{fare} \textit{\textbf{schifo}}\\ do.\textsc{inf} disgust.\textsc{sg}\\ ~ noun.feeling\\
    \ex \label{ex:truetag2}
    \glll\textit{fare} \textit{\textbf{paura}}\\ do.\textsc{inf} fear.\textsc{sg}\\ ~ noun.feeling\\
    \ex \label{ex:truetag3}
    \glll\textit{fare} \textit{\textbf{piacere}}\\ do.\textsc{inf} pleasure.\textsc{sg}\\ ~ noun.feeling\\
    \end{xlist}
\end{exe}
\begin{exe}
\ex \label{ex:falsetag}False positives:
    \begin{xlist}
       \ex \label{ex:falsetag1}
    \glll\textit{fare} \textit{\textbf{demagogia}}\\ do.\textsc{inf} demagogy.\textsc{sg}\\ ~ noun.communication\\
           \ex \label{ex:falsetag2}
    \glll\textit{fare} \textit{\textbf{parte}}\\ do.\textsc{inf} part.\textsc{sg}\\ ~ noun.group\\
           \ex \label{ex:falsetag3}
    \glll\textit{fare} \textit{\textbf{cassa}}\\ do.\textsc{inf} cash.\textsc{sg}\\ ~ noun.quantity\\
    \end{xlist}
\end{exe}





\subsection{Coverage of Italian Treebanks lexicon}

Since the primary aim of our formalization is to map cxns in ItCon to UD-annotated corpora, as a preliminary evaluation of our tagset we checked how many lemmas and how many forms in Italian \lp{UD} treebanks are associated to at least one synset (and thus, at least one OMW topic) in Italian MultiWordNet. 
We extracted the frequency lists for noun and verb lemmas from Italian treebanks\footnote{\url{https://universaldependencies.org/\#italian-treebanks} with the exception of Italian-Old~\cite{corbetta2023highway}, as it is actually a treebank of old Italian, containing Dante Alighieri's \LB{\textit{Divine Comedy}.}}, and selected the lemmas with frequency higher than 5 (n = 5273). We then extracted all the synsets and the associated \textit{lexnames} (OMW topics) for each lemma, using \texttt{NLTK}\footnote{\url{https://www.nltk.org/}} WordNet interface in Python to access data from \texttt{omw-it 1.4}.

Though not all the lemmas in Italian Treebanks have a corresponding OMW topic, the results are encouraging (\lp{Appendix \ref{sec:coverage}}
). Only 10\% of the noun lemmas (n = 394) and 12.7\% of the verb lemmas (n = 173) are not assigned any semantic tag\bb{s} (Table \ref{tab:cover_lemma}). Moreover, the percentage of untagged noun\bb{s} and verb\bb{s} in Italian Treebank gets lower if we look at the forms count (obtained by adding together the frequencies of the lemmas). Namely, only 3.5\% of the forms (both for the verbs and for the nouns) is not associated to any topic\bb{s} (Table \ref{tab:cover_form}).

Although a broader coverage of the Italian treebanks would be desirable, also considering that we \LB{set a strict frequency threshold, these results are promising. In fact, they suggest that a substantial number of constructs can be identified using our semantic annotation.}

\subsection{Limitations}

Nonetheless, using OMW topics as semantic tags can bear some limitations. 
\LB{Firstly,} a pre-defined classification \LB{does not necessarily include all needed semantic classes}, as opposed to a bottom-up classification\footnote{\fm{See for instance the approach taken in \citet{jezek2014t}.}}: \LB{using an existing widely used ontology makes adding new, \textit{ad hoc} semantic tags impossible, as it would hinder interoperability with other WordNet-connected resources.}
\LB{Secondly}, the semantic classification is only available for nouns and verbs, since there are no top nodes for adverbs and only three top nodes for adjectives (\texttt{all}, \texttt{participial}, \texttt{pertainyms}). Currently, the choice of the semantic tagset for adjectives and adverbs stands as an open challenge: while at least for adjectives some classifications exist (e.g., \citealt{dixon-2004}), also in the context of some WordNets (e.g., GermaNet, \citealt{germanet}), they are not mapped onto Italian resources. Thus, while they could be used for descriptive purposes, it would be difficult to employ them consistently in the matching process.

\section{Future steps: annotation of inter-slot semantic relations}
\label{sec:future}

A challenge \LB{for} our formalization is \LB{represented} by the cases in which constraining the fillers of a single slot is not enough in order to match the instances of a cxn.
As a matter of fact, the idiosyncratic behaviour of some syntactic and multiword cxns consists in the semantic interdependence of their slots~\citep{desagulier2016}. Some examples include:

\begin{exe}
	\ex \label{ex:oxy} Oxymorons \jambox{\citep{la-pietra-masini}}
	\begin{xlist}
		\ex \label{ex:oxy1} \glll \textit{l'} \textit{ingiustizia} \textit{della} \textit{giustizia}\\
  \textsc{det.f.sg} injustice.\textsc{sg}  of.\textsc{det.f.sg} justice.\textsc{sg} \\ 
  ~ noun.attribute ~ noun.attribute\\
		\glt `the injustice of justice'
		\ex \label{ex:oxy2} \glll \textit{allegria} \textit{triste}\\ 
  joy.\textsc{sg} sad.\textsc{sg}\\ 
  noun.feeling adj.all\\
		\glt `sad joy'
        \end{xlist}
    \ex \label{ex:co} Cognate cxns \jambox{\citep{MelloniMasini2017, Busso2020}}
	\begin{xlist}
		\ex \label{ex:co1} \glll \textit{vivere} \textit{la} \textit{vita}\\ 
  live\lp{.\textsc{inf}} \textsc{det.f.sg} life.\textsc{sg}\\ 
  verb.stative ~ noun.state\\
		\glt `to live life'
		\ex \label{ex:co2} \glll \textit{danzare} \textit{una} \textit{danza}\\ 
  dance\lp{.\textsc{inf}} \textsc{det.f.sg} dance\textsc{.sg}\\ 
  verb.motion ~ noun.act\\
		\glt `to dance a dance'
        \end{xlist}
\end{exe}
        
For instance, in (\ref{ex:oxy}) the two slots are filled by antonymic words, while in (\ref{ex:co}) the verb and the object are derivationally or semantically related. In such cxns, acknowledging the paradigmatic or semantic relationship between the fillers is necessary in order to define the cxns and to distinguish such instances from other formally similar cxns. Such relations can take place between same-POS fillers (\ref{ex:oxy1}) but also between different-POS fillers (\ref{ex:oxy2}, \ref{ex:co}).

A possible solution could be to \LB{use} the network structure of WordNet. As a matter of fact, 
OMW topics are only taken as a semantic classification, since they are top nodes of the hierarchy and no semantic relation is specified among them (let alone cross-POS relations). It should \LB{therefore} be quite straightforward to constrain the possible fillers by checking \bb{if a specific semantic relation between two fillers exists in WordNet's database}. This can be implemented through the \texttt{IDENTITY} field in the CoNLL-C formalization.

Normally, we employ the \texttt{IDENTITY} field to specify if two or more fillers' fields should have the same value \lp{for a given feature}. For instance, Table \ref{tab:NnotN} shows how \bb{this} field is employed in the case of discontinuous reduplication cxns N\textsubscript{i} \textit{non} N\textsubscript{i} `N not N' \fm{(e.g. \textit{sapone non sapone}, lit. soap not soap, meaning `soap-free detergent')} \cite{masini-di-donato}.

\begin{table}[ht]
{\small
\begin{tabular}{cccccc}
\hline
\texttt{ID} & \texttt{UD.FORM} & \texttt{LEMMA} & \texttt{UPOS} & ... & \texttt{IDENTITY} \\
\hline
A  & \_       & \_  & NOUN              & ... &  \_          \\
B  & non       & non     & ADV    & ... &  \_  \\ 
C  &  \_  & \_  & NOUN     & ... &  UD.FORM=A \\
\hline
\end{tabular}}
  \caption{
    Partial CoNLL-C formalization of N\textsubscript{i} \textit{non} N\textsubscript{i} `N not N' cxn.
  }
  \label{tab:NnotN}
\end{table}



\fp{However, \texttt{IDENTITY} can easily be adapted to represent same-POS relations by making reference to WordNet relations between synsets: }
for instance, in a oxymoronic N\textsubscript{1} Prep N\textsubscript{2} cxn such as (\ref{ex:oxy1})\bb{, }the synsets of the two nouns are linked by an \texttt{antonym} relation. This relation could be formalized, so as to remain queryable in WordNet, as:

\begin{lstlisting}
LEMMA=antonym:N$_1$    
\end{lstlisting}


Problems arise in case of different-POS fillers, such as in \textit{Cognate Object cxns}, where the verb and the object are semantically (and often derivationally) related, the object being a shadow argument of the verb.
While MultiWordNet does not encompass cross-POS relations, ItalWordNet~\citep{italwordnet2000} includes a number of cross-POS relations, inherited from EuroWordNet~\citep{eurown}\footnote{See~\citet{Alonge1998} and~\citet{italwordnet2000} for a description of the relations.}. 


However, by consulting the most recent OMW-compliant version of ItalWordNet\footnote{\url{https://github.com/valeq/IWN-OMW/}}~\citep{quochi-etal}, such relations seem to be \fp{employed }only partially. Nonetheless, the behaviour of the constructs in (\ref{ex:co}) is captured in ItalWordNet: \textit{danzare} `to dance' is in a \texttt{similar} relation with \textit{danza} `dance', pointing that the two synsets express similar meanings\footnote{Actually, the \texttt{similar} relation was not defined in EuroWordNet, but is part of the Princeton WordNet relations (\url{https://globalwordnet.github.io/schemas/\#rdf}).}, and the same holds for \textit{vivere} `to live' and \textit{vita} `life'. 

While being ideally very powerful for our formalization, such \bb{an }approach needs a wide and consistent coverage of the Italian lexicon and its relations in WordNet. This is needed in order to avoid filtering out possible instances of the cxns if a semantic relation is absent in WordNet. For instance, quite common examples of Cognate cxns (\ref{ex:notfound1}-\ref{ex:notfound2}) would be filtered out since the verb and the object bear no relation in ItalWordNet:
\begin{exe}
\ex \label{ex:notfound1} {\gll \textit{Sara} \textit{\textbf{ha}} \textit{\textbf{dormito}} \textit{un} \textit{\textbf{sonno}} \textit{di} \textit{piombo}\\ 
Sara \textsc{aux.3sg} sleep.\textsc{pst}   \textsc{det.m.sg} sleep.\textsc{sg} of plumber.\textsc{sg}\\
\glt`Sara slept a deep sleep.'}\jambox{\cite{Busso2020}}
\ex \label{ex:notfound2} {\gll \textit{\textbf{Ho}} \textit{\textbf{sognato}} \textit{un} \textit{bel} \textit{\textbf{sogno}} \textit{stanotte.}\\
\textsc{aux.1sg} dream.\textsc{pst} \textsc{det.m.sg} beautiful.\textsc{m.sg} dream.\textsc{sg} last\_night\\
\glt`Last night, I dreamed a beautiful dream.'} \jambox{(adapted from \citealt{MelloniMasini2017})}
\end{exe}

Nonetheless, for the moment this annotation can still be useful at the descriptive level, since it provides us \bb{with a consistent way to annotate constructional properties of our entries in a fine-grained fashion}, and will be hopefully exploited in the future for the matching process.

\section{Conclusions}
The present contribution \bb{has outlined} how the Italian Constructicon project aims at making lexical and constructional resources interoperable in a fruitful manner. We have shown how WordNet's network structure can be \fm{employed} 
to flexibly describe the idiosyncratic behaviour of constructions.
The biggest limitations of this approach are practical in nature. 
In fact, this protocol would \bb{work properly only with} 
a greater coverage of OMW semantic classification, together with Italian corpora annotated with (super)senses. 
\fp{Moreover, as ItCon  is committed to include morphological (i.e., word-formation) cxns
, an 
\LB{unanswered} question is whether this semantic classification will prove to be adequate for the annotation of semantic constraints in morphological cxns as well (as for now it has been employed for multiword and syntactic constructions).}
Despite these open questions, and despite the ItCon project \LB{still} being in its infancy, we have \fm{shown} 
how using Open Multilingual WordNet to represent cxns' semantic features is a fruitful way to link different types of language resources, making them interoperable cross-linguistically.




\bibliography{custom}

\appendix

\section{Coverage}
\label{sec:coverage}

\begin{table}[!ht]
\fontsize{8.5pt}{8.5pt}\selectfont

    \begin{tabular}{lcc}
    \hline
        \textbf{class} & \textbf{n. lemmas} & \textbf{n. forms} \\ \hline
        noun.Tops & 58 & 9153 \\ 
        noun.artifact & 1744 & 31035 \\ 
        noun.act & 1566 & 45486 \\ 
        noun.person & 1338 & 22933 \\ 
        noun.communication & 1211 & 40686 \\ 
        noun.attribute & 862 & 25920 \\ 
        noun.cognition & 805 & 35952 \\ 
        noun.state & 714 & 24911 \\ 
        noun.group & 525 & 30015 \\ 
        noun.event & 366 & 9797 \\ 
        noun.substance & 279 & 3809 \\ 
        noun.location & 267 & 13602 \\ 
        noun.possession & 261 & 12188 \\ 
        noun.animal & 251 & 3457 \\ 
        noun.object & 237 & 5300 \\ 
        noun.feeling & 235 & 4024 \\ 
        noun.body & 231 & 6398 \\ 
        noun.quantity & 215 & 7920 \\ 
        noun.food & 208 & 1651 \\ 
        noun.time & 203 & 14581 \\ 
        noun.plant & 200 & 1559 \\ 
        noun.phenomenon & 141 & 5869 \\ 
        noun.relation & 116 & 5561 \\ 
        noun.process & 112 & 3314 \\ 
        noun.shape & 89 & 2408 \\ 
        noun.motive & 20 & 1329 \\ 
        verb.change & 552 & 16761 \\ 
        verb.communication & 550 & 20294 \\ 
        verb.contact & 477 & 11470 \\ 
        verb.social & 414 & 15671 \\ 
        verb.motion & 291 & 10265 \\ 
        verb.cognition & 273 & 12995 \\ 
        verb.possession & 260 & 11749 \\ 
        verb.stative & 259 & 15682 \\ 
        verb.creation & 210 & 8864 \\ 
        verb.body & 168 & 4437 \\ 
        verb.emotion & 158 & 3547 \\ 
        verb.competition & 119 & 4594 \\ 
        verb.consumption & 80 & 4099 \\ 
        verb.perception & 27 & 6782 \\ 
        verb.weather & 27 & 397 \\
        \hline
    \end{tabular}
     \caption{
     Count of lemmas and forms (nouns and verbs only) for each OMW topic (a lemma can belong to more than one topic).}
\label{tab:n_nouns}
\end{table}

\begin{sidewaystable} 
\centering
	\small{
	\begin{adjustbox}{scale=0.95,center}
	\begin{tabular}{ccccccccccccc}
			\hline
        \textbf{POS} & \textbf{0} & \textbf{1} & \textbf{2} & \textbf{3} & \textbf{4}  & \textbf{5}  & \textbf{6}  & \textbf{7}  & \textbf{8}  & \textbf{9}  & \textbf{11}  & \textbf{Total} \\ \hline
		\textit{noun} & 394 (10,1\%)  & 1882 (48,2\%)  & 913 (23,4\%)  & 429 (11,0\%)  & 172 (4,4\%)  & 83 (2,1\%)  & 18 (0,5\%)  & 9 (0,2\%)  & 4 (0,1\%)  & 3 (0,1\%)  & 0 (0\%)  & 3907 (100,0\%)  \\ 
		\textit{verb} & 173 (12,7\%)  & 503 (36,8\%)  & 369 (27,0\%)  & 159 (11,6\%)  & 89 (6,5\%)  & 43 (3,2\%)  & 20 (1,5\%)  & 4 (0,3\%)  & 3 (0,2\%)  & 1 (0,1\%)  & 2 (0,2\%)  & 1366 (100,0\%)  \\ \hline
		\textbf{Total} & 567 (10,8\%)  & 2385 (45,2\%)  & 1282 (24,3\%)  & 588 (11,2\%)  & 261 (5,0\%)  & 126 (2,4\%)  & 38 (0,7\%)  & 13 (0,3\%)  & 7 (0,1\%)  & 4 (0,1\%)  & 2 (0,0\%)  & 5273 (100,0\%) \\ \hline
	\end{tabular} \end{adjustbox} }
 \caption{Counts and percentages and  of noun and verb lemmas by number of OMW topics in Italian Treebanks.}
\label{tab:cover_lemma}

 \vspace{1in}

\small{
	\begin{adjustbox}{scale=0.95,center}
			\begin{tabular}{ccccccccccccc}
				\hline
        \textbf{POS} & \textbf{0} & \textbf{1} & \textbf{2} & \textbf{3} & \textbf{4}  & \textbf{5}  & \textbf{6}  & \textbf{7}  & \textbf{8}  & \textbf{9}  & \textbf{11}  & \textbf{Total} \\ \hline
\textit{noun} & 5388 (3,5\%)  & 49610 (31,9\%)  & 40024 (25,8\%)  & 23757 (15,3\%)  & 18327 (11,8\%)  & 12270 (7,9\%)  & 2263 (1,5\%)  & 1340 (0,9\%)  & 850 (0,6\%)  & 1483 (1,0\%)  & 0 (0\%)  & 155312 (100,0\%)  \\
\textit{verb} & 2449 (3,5\%)  & 14439 (20,4\%)  & 15196 (21,4\%)  & 12276 (17,3\%)  & 8663 (12,2\%)  & 4161 (5,9\%)  & 3529 (5,0\%)  & 1197 (1,7\%)  & 2995 (4,2\%)  & 416 (0,6\%)  & 5595 (7,9\%)  & 70916 (100,0\%)  \\ \hline
\textbf{Total} & 7837 (3,5\%)  & 64049 (28,3\%)  & 55220 (24,4\%)  & 36033 (15,9\%)  & 26990 (11,9\%)  & 16431 (7,3\%)  & 5792 (2,6\%)  & 2537 (1,1\%)  & 3845 (1,7\%)  & 1899 (0,8\%)  & 5595 (2,5\%)  & 226228 (100,0\%) \\ \hline
\end{tabular} \end{adjustbox} }  
 \caption{Counts and percentages of noun and verb forms by number of OMW topics in Italian Treebanks.}
\label{tab:cover_form}
\end{sidewaystable}

\end{document}